IAC-24-A3.IP.96.X84000

# Towards Safer Planetary Exploration: A Hybrid Architecture for Terrain Traversability Analysis in Mars Rovers

Achille Chiuchiarelli[a]*, Giacomo Franchini[a], Francesco Messina[a], Marcello Chiaberge[a]

[a] *Department of Electronics and Telecommunications - PoliTo Interdepartmental Center for Service Robotics (PIC4SeR), Polytechnic of Turin, Corso Duca degli Abruzzi, 24    10129 Torino, Italy,*
*s303215@studenti.polito.it, giacomo.franchini@polito.it, francesco_messina@polito.it, marcello.chiaberge@polito.it*
\* Corresponding Author

**Abstract**

The field of autonomous navigation for unmanned ground vehicles (UGVs) is in continuous growth and increasing levels of autonomy have been reached in the last few years. However, the task becomes more challenging when the focus is on the exploration of planet surfaces such as Mars. In those situations, UGVs are forced to navigate through unstable and rugged terrains which, inevitably, open the vehicle to more hazards, accidents, and, in extreme cases, complete mission failure. This paper tackles the problem of terrain traversability analysis in the context of planetary exploration rovers, delving particularly into Mars exploration. The research aims to develop a hybrid architecture, which enables the assessment of terrain traversability based on the results of both an appearance-based approach and a geometry-based approach. The coexistence of the two methodologies has the objective of balancing each other's flaws, reaching a more robust and complete understanding of the operating environment. The appearance-based method employs semantic segmentation, operated by a deep neural network, to understand the different terrain classes present in the scene. Predictions are refined by an additional module that performs pixel-level terrain roughness classification from the same RGB image. The rationale behind this choice resides in the will to assign different costs, even to areas belonging to the same terrain class, while including an analysis of the physical properties of the soil. This first cost map is then combined with a second one yielded by the geometry-based approach. This module evaluates the geometrical characteristics of the surrounding environment, highlighting categories of hazards that are not easily detectable by semantic segmentation. The proposed architecture has been trained using synthetic datasets and developed as a ROS2 application to be easily integrated into a higher-level framework for autonomous navigation in harsh environments. Simulations have been performed in Unity, showing the ability of the method to assess online traversability analysis.

**Acronyms/Abbreviations**
Unmanned Ground Vehicle (UGV),
Red Green Blue Depth (RGBD),
JPL (Jet Propulsion Laboratory),
MSL (Mars Science Laboratory),
MER (Mars Exploration Rover),
NAVCAM (Navigation Camera),
MASTCAM (Mast Camera),
PBR (Physically Based Rendering)

## 1. Introduction

Space exploration has been one of the biggest goals of humankind, starting from the 50s to the present day, pushing further and further the boundaries of what it is possible to achieve. Among the many routes through which this field branches, a topic always under the spotlight is planetary exploration. Beginning from the race to the Moon, the will to explore celestial bodies, different from the Earth, has never faded away and, after the latter has been conquered, the focus rapidly moved towards the red planet, Mars. It is exactly in this context that this research puts its bases, presenting a possible framework for the terrain traversability analysis of a planetary exploration rover autonomously navigating through the Martian surface.

This process features the study of the environment around the rover to gain an understanding of the surroundings. This finalizes the achievement of safe and aware navigation without the need for human intervention. The information retrieved during this procedure is essential in the avoidance of hazards that could damage the vehicle and endanger the correct progress of the mission. The data can be acquired from several sources and represent diverse characteristics of the terrain.

This research will present a hybrid approach, combining an appearance-based and a geometry-based approach to assess online terrain traversability analysis. This paper contributes a novel hybrid architecture employing only RGBD data to evaluate the environment both visually and geometrically, with the addition of a refining measure studying also roughness, a physical property of the terrain.

This paper aims to develop a framework able to assess, from the information coming from a single sensor, online terrain traversability in an unstructured environment to be employed in an autonomous navigation system.

*1.1 Literature Review*

The use of UGVs in planetary exploration is fundamental. Since space exploration is characterized by extreme conditions, high risks, really dilated times, and enormous costs, it is still not suitable for a growing presence of human operators, and it finds a great aid in the employment of rovers.

It is possible to identify three common problems that need to be addressed to make rovers' operations feasible [1]:

- Assessment of terrain traversability.
- Planning optimal motion paths concerning given criteria.
- Suitable adaptation of the kinematic configuration of articulated robots as a function of terrain traversability.

Even if the three topics are highly interdependent, due to their complexity they are usually studied individually. In this paper, the research focuses particularly on the first one, terrain traversability analysis. When discussing the approach to evaluate terrain traversability, it is possible to divide them into two categories based on the source from which the data are gathered: *Proprioceptive* approaches, using internal rover's information, and *Exteroceptive* approaches, taking advantage of data coming from the environment surrounding the vehicle. The latter can be further divided into *Appearance-based* approaches and *Geometry-based* approaches.

The first one includes all the image processing operations with either the aim of classification or regression of terrain characteristics [2], [3], [4], while the second one revolves around the capability of the UGV to estimate the geometric characteristics of the environment surrounding it [5], [6], [7].

While both sides have been previously employed successfully to singlehandedly assess terrain traversability, they also present some significant flaws that could endanger the navigation of the rover. To try to compensate them the research landscape moved in the direction of *Hybrid* approaches, combining different methods to increase their robustness. In some of the earlier research on the theme, like [8] and [9], the approach was to use the geometric portion mainly for obstacle detection while the appearance side to provide additional information. Another interesting procedure is presented in [10], where the two data sources are combined to be the input of a fully convolutional network classifying every pixel directly based on its traversability.

The methodology employed in this paper is, instead, pursuing a more balanced use of the two methods similarly to [11]. Both sides are employed to compensate for each other's drawbacks combining the traversability cost maps they create into a final one, considering a wider range of features.

*1.2 Research' Structure*

The remainder of this paper is organized as follows. Section 2 is devoted to the methodologies employed in the development of the traversability framework, describing its overall architecture and detailing every component of it. Section 3 describes the datasets of Martian images employed, explaining the rationale behind the choices made in its definition and the process leading to the final version of it. Section 4 elaborates on all the key development phases and experiments chosen to test the effectiveness and performance of the framework. Section 5 contains the results of the testing phases and the conclusions that can be drawn from them. In the end, in Section 6, the presented research is summarized and a brief discussion on some proposed future improvements is reported.

**2. Methods**

The proposed framework follows an architecture like the one presented in [11], an appearance-based approach is combined with a geometry-based approach through a weighted sum of the relative cost maps to get to the final hybrid result.

*2.1 Appearance-based approach*

The appearance-based side employs two modules: a first one applying semantic segmentation of the 4 most relevant Martian terrain classes and a second one predicting the terrain's roughness, solely from visual inputs.

*2.1.1 Semantic Segmentation module*

In [12] NASA's JPL laboratory defines these 4 main classes to be *Soil*, *Bedrock*, *Sand*, and *Big rocks*, with the last two representing the most hazardous ones for the navigation of UGVs.

Semantic segmentation is performed using a deep fully convolutional neural network, UNet [13], trained employing a dataset of synthetic Martian terrain images. RGB images originally captured with a resolution of 1024x1024 are downscaled to 256x256 to facilitate computational efficiency and conform to the network's input dimensions.

The module has been trained over a set of 670 images split into training, validation, and test sets.

Due to the imbalance concerning the *Big rock* class, efforts are made to compensate for it using Weighted Cross Entropy as the loss function. It follows the same working principles as the standard one, but considers a weight for every class, chosen to be proportional to the inverse of the classes' percentages in the dataset. The network is trained for 15 epochs, with a starting LR of $10^{-3}$ decreased by 30% every two epochs through a learning rate scheduler. Adam Optimizer [14] is employed.

The segmentation map, created by the trained network instance, is used to obtain its relative cost map. To each class, a cost reduction, ranging between 0 and 1, is empirically assigned, based on the rover's structure and previous exploration missions' failures.

Table 1. Semantic segmentation cost reductions

|  | Soil | Bedrock | Sand | Big rock |
|---|---|---|---|---|
| Reduction | 0.8 | 0.7 | 0.5 | 0.0 |

The values displayed in the table above are multiplied by the predictions' confidence to follow a more cautious approach.

*2.1.2 Roughness Classification module*

To increase the detail level of the appearance-based result, an additional module has been introduced to predict, from RGB images, roughness levels of the terrain. Starting from the proposal formulated in [4], a module in charge of roughness prediction from terrain images has been implemented. Another instance of UNet is used to classify four different roughness levels, defined a priori based on the risk level represented to the rover's traversability. The roughness levels range from *LV0 (Negligible roughness)*, between 1 and 4 mm, to LV3 *(High roughness)*, for every pixel above 10 cm, most likely belonging to an obstacle.

The training process of this additional module follows the same setup employed for semantic segmentation. The network undergoes 15 epochs of training and Weighted Cross Entropy is, again, chosen as the loss function given the even more unbalanced nature of the roughness dataset. The set is composed of the same 670 RGB images, with custom-made roughness ground truths. The dataset creation and processing will be detailed, as mentioned previously, in section 3.

The same approach to build the semantic segmentation cost map is applied to obtain the roughness classification one. Again, empirical cost reductions, based on the potential risk to the navigation of the rover, are assigned to each class, as can be seen in the following table.

Table 2. Roughness classification cost reductions

|  | LV0 | LV1 | LV2 | LV3 |
|---|---|---|---|---|
| Reduction | 0.85 | 0.6 | 0.4 | 0.1 |

As for the previous case, predictions' confidence is accounted for in the creation of the cost map.

*2.1.3 Appearance-based cost map*

The two maps are subsequently combined through a weighted sum process to obtain the final appearance-based cost map. As anticipated, the rationale behind the introduction of the roughness module resides in the will to increase the detail level of the map, being able to differentiate among regions belonging to the same class. Accordingly, the corresponding weight will be lower than the one assigned to the semantic segmentation result, which is used as a baseline for the final map.

*2.2 Geometry-based approach*

The other side of the framework oversees the assessment of terrain traversability from a geometrical perspective. This module starts from the same RGB samples evaluated by the appearance-based side too, but employs also the corresponding depth data, which could come from either an RGBD camera or a couple of stereo cameras, to estimate the terrain's slope in every point. These data are handled by the *Python library Open3D* [15], to generate the point cloud corresponding to the environment depicted in the acquired picture. From the cloud, a series of transformations is needed to get the data in the correct reference frame.

These operations are required due to the desire to generate a robot-centric cost map, concerning the hybrid result, so a map centered around the robot's base-link with the three axes following the rover's motion.

Once the robot-based point cloud is obtained, the normal vectors for each point are estimated by calculating the principal axis of the adjacent points using covariance analysis. Due to covariance analysis producing two opposite directions for the normal vectors, they must be guided in some way by the user to select only the ones outgoing from the point cloud surface. To do it, a method of the Open3D library has been employed to orient them according to an orientation reference, in this case, the positive vertical axis. From these vectors, the angles between them and the vertical axis are evaluated to represent the terrain's slope at each point. The slope values are the base for the assignment of the geometry-based cost to be inserted in the relative cost map.

The cost assignment process is regulated by a custom piecewise function reported in Equation 1. The function is made by linear functions with different slopes combined and saturated to the maximum cost value of 1 inside certain specific angles' intervals. Two variable values, called $t_{soft}$ (soft threshold) and t (threshold) are defined to switch among the different portions of the function. The first one, whose value is selected to 30°, marks the transition from an initially gentler increase in the costs' values to a second interval characterized by a steeper rise. The latter is defined to emphasize the risk represented by higher slopes and ends with the maximum cost value in correspondence to the threshold one, chosen at 70°. This value is the starting point of the interval of untraversable slopes. The same assignment is done considering opposite inclinations of the terrain, which could verify when the vehicle navigates down a hill or is in the presence of negative geometrical obstacles.

$$cost_{geom}(\theta_p) = \begin{cases} \frac{2}{\pi} \cdot \theta_p, & \text{if } \theta_p < t_{soft} \\ \frac{1}{t} \cdot \theta_p, & \text{if } t_{soft} \leq \theta_p \leq t \\ 1, & \text{if } t < \theta_p < 2\pi - t \\ 1 - \frac{[\theta_p - (2\pi - t)]}{t}, & \text{if } 2\pi - t \leq \theta_p \leq 2\pi - t_{soft} \\ 1 - \frac{\theta_p - \frac{3\pi}{2}}{\frac{\pi}{2}}, & \text{if } 2\pi - t_{soft} < \theta_p \leq 2\pi \end{cases}$$

(1)

The resulting values are employed, in the end, to fill a robot-centric grid map.

*2.2 Cost maps combination*
The products of the two sides of the framework are combined to yield the final hybrid traversability map. To facilitate this merging process the appearance-based map, initially represented in the image plane, needs to be re-projected into the correct reference frame. To perform this transformation a characteristic of the Open3D library is exploited, the order of the unstructured point cloud obtained from a certain picture reflects the same of the image pixels. Then, knowing the correspondence between each point and the grid cells, defined during the creation of the geometry-based cost map, and the correspondence between each pixel and each point it is easy to find the one between pixels and grid cells, obtaining the projected version from the image plane to the robot-base frame.

Once this preliminary processing phase is completed, the two maps can finally be combined through a weighted sum. The hybrid traversability cost of each cell is therefore obtained as:

$$c_H = w_A \cdot c_A + w_G \cdot c_G \qquad (2)$$

**3. Dataset**
When working with deep learning vision systems, like it is done for semantic segmentation and roughness classification, another requirement arises, the need for a large-scale set. Being this paper focused on the domain of planetary exploration, it was needed to find a set of images and, more importantly, of labels, coming from a reliable source and validated by scientific experts.

These same demands led to the creation of AI4Mars [12], the first large-scale dataset for training and validation of terrain classification models for Mars. The size of this set is particularly highlighted due to the effect it can have on the performances of deep learning models. The dimension of the dataset plays a crucial role relative to the learning capabilities of the network. A large and diverse dataset will enhance the generalization skills towards unseen data, improving the robustness of the training and having, moreover, potential benefits towards overfitting.

*3.1 AI4Mars*
To try to actively employ vision systems in the autonomous navigation of Martian rovers, researchers at JPL, in [2], proposed a new machine learning-based terrain classifier for Mars named SPOC (Soil Property and Object Classification). It uses a deep convolutional network to identify terrain types and terrain features. The results obtained were encouraging, but to be able to obtain the necessary reliability levels for on-board algorithms' standards a high-quality dataset was needed.

This dataset includes the majority of existing high-resolution images of Mars' surface. It gathers samples from several UGVs from MSL and MER missions taken with both NAVCAM and MASTCAM. The set, as previously mentioned, is split into four different label categories: Soil, Bedrock, Sand, and Big rocks.

Even if it embodies all the characteristics required to train successfully the semantic segmentation module, AI4Mars lacks a portion of data fundamental for the training of the roughness classification network, since to create the relative ground truth each RGB image must be coupled with the corresponding depth mask.

*3.2 Synthetic dataset*
To overcome the lack of information, a custom synthetic dataset has been created. The images are generated through the use of a photo-realistic simulator called *Oaisys* [16]. Oaisys (Outdoor Artificial Intelligent

SYstems Simulator), has been created to satisfy the open demand for high-quality synthetic data for planetary exploration tasks and to allow all the detailed operations required in the generation of those data. Among the most important features provided by the software, it is possible to include the capability of parametric development of the entire environment and the generation of high-fidelity metadata. Condensing the working principles of the simulator, it generates diverse landscapes by deforming a base mesh, called the stage, considered as the ground layer. Materials are defined through PBR textures and applied to the stage; they can also be combined via a noise shader to create more realistic results. Each texture is coupled with a corresponding semantic ID to easily retrieve semantic information. Objects, defined from blender-based meshes with appropriate textures, are scattered in the environment using noise maps. Lighting is configured by the simulator by either selecting custom HDR images at random or employing Blender's Sky simulator. Camera poses can be determined sequentially from a *csv* file or randomly from predefined intervals, specified for both the sensor and its target.

Adjusting the configuration file to resemble as much as possible the Martian landscape, the synthetic dataset has been created. The generation of these artificial images presents the advantages of having the opportunity to customize a lot of parameters, the pose of the camera, the texture used for each material, but more importantly to couple the RGB data with the information needed. In this way, it has been possible to obtain a series of samples correlated with highly accurate semantic masks and relative depth maps.

*3.2.1 Roughness ground truths*

To train the roughness classification module a custom dataset of roughness ground truths is required. Being not directly retrievable from Oaisys they must be created specifically. As mentioned previously, the information obtained from the simulator is limited to the depth maps retrieved for every image, the next step is the generation, from the available data, of equivalent roughness maps. The method presented in [4] has been followed with some adjustments.

It can be summarized in four steps:
1. Point cloud creation, differently from the approach described in the aforementioned research, the cloud is obtained from the Python library Open3D. It is created starting from the depth map and RGB image, to use as much data as possible. The final result is achieved in two steps, at first an RGBD image is created combining the available data and it is, then, used to generate the point cloud.
2. Plane fitting, deviating again from the considered paper, in this case, the plane fitting procedure has been applied over limited regions of the image by dividing it into a series of patches. To capture as many roughness variations as possible, the patches' dimension is modified, depending on the depth values, considering bigger patches as closer areas are considered. For each patch, a corresponding point cloud is created, and plane fitting is applied, trying to find the coefficients of the plane that best approximate the studied surface.
3. Roughness computation, once the plane coefficients for a patch have been obtained, the following formula for the evaluation of the roughness level of each point, presented in [4], is employed to retrieve the corresponding value.

$$r_i = \frac{|-d-a\cdot x_i - b\cdot y_i - c\cdot z_i|}{\sqrt{a^2+b^2+c^2}} \qquad (3)$$

4. Roughness map definition, the roughness values computed in the previous step, are assigned to the relative pixels in the image making use of the correspondence between the points' indexes, in the unstructured point cloud, which can be retrieved using a method of Open3D, and the indexes of the pixels in the 2D NumPy array defining the map. Post-processing operations are done on the obtained map cutting the roughness values at the mm scale, imposing every lower roughness value to be equal to 1mm, and assigning each pixel to the corresponding roughness class.

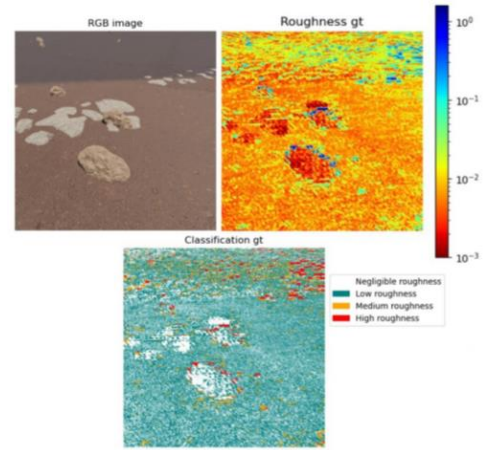

Fig 1. Sample of a roughness ground truth

## 4. Development

The proposed architecture is implemented as a ROS2 application to be integrated easily into a higher-level framework for autonomous navigation in harsh environments. It is developed as a Python package comprehending 3 different nodes which will be started at once thanks to a launch file. They create the complete data flow which from the input data, coming from the rover, leads to the output of the final traversability cost map.

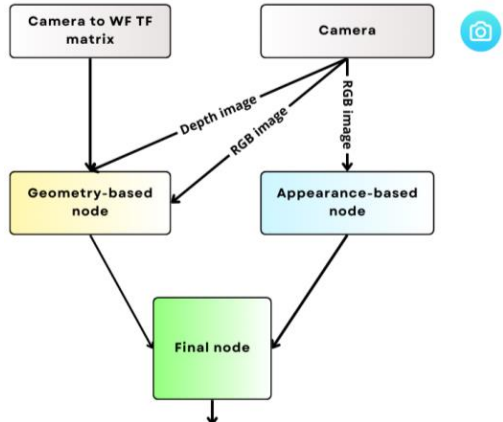

Fig 2. Complete ROS2 architecture

The information, coming from the camera on the UGV, will be sent to the two nodes in charge of the hybrid traversability analysis which will process them creating the two, already discussed, traversability maps. The outputs of the nodes will go through an approximate time synchronizer, inside the final node, before being employed in the creation of the hybrid cost map. This last node will oversee the remaining processing operations on the maps and their combination, to get to the final desired result.

*4.1 Simulation tests*

To verify the effectiveness of the proposed framework to be employable in an autonomous navigation system, simulation tests have been performed to evaluate its performance in simulated working conditions. The final goal of these experiments is to achieve the possibility of safe navigation of a simulated rover in a Martian-like environment based solely on the traversability evaluations of the presented architecture.

*4.1.1 Simulation setup*

The setup for the aforementioned trials involves the creation of a simulated Martian landscape in which the vehicle can navigate. To serve this purpose it has been decided to employ Unity, a cross-platform game engine widely used in the creation of both 3D and 2D games. It is extensively adopted for activities outside the gaming industry to create interactive simulations. Among the many fields involved, it is possible to find film, automotive, architecture, engineering, construction industries, and even the United States Armed Forces.

In this specific case, the creation of the environment started from the output *blend* file created by Oaisys itself. From this baseline, the simulation has been reviewed and adjusted, mainly by reorganizing the assets displaced over the terrain mesh and simplifying them to lighten the file to be processed by Unity. To export it from Blender into the new environment maintaining the same result, a baking procedure of the terrain texture was needed, being it the result of the mix between two different materials through a noise map.

Once the preprocessing operations have been completed, the *fbx* file has been imported into Unity. All the materials involved have been restored to obtain a result as resembling as possible to the images used in the networks' training.

In the simulation, the rover has been added and equipped with the required sensors to capture all the data essential for assessing traversability.

The next phase involved the linking procedure between the ROS2 nodes and the Unity environment. It has been performed using the *ros_tcp_endpoint* package [17], a package used to create an endpoint to accept ROS messages sent from a Unity scene. The rover is, then, moved inside the scene recording all the acquired data, published through the relative topics, to create a rosbag. The result is employed to test the framework's effectiveness in evaluating terrain traversability.

The previously presented nodes oversee the analysis of the environment depicted in the images and assess traversability creating the hybrid cost map.

The map is used, in the end, as a base to plan the rover's motion employing a path-planning algorithm.

*4.1.2 Experiments*

The experiments carried out in the simulation environment revolved, as anticipated, all around the assessment of the framework's effectiveness in simulated working conditions. The Rover is teleoperated inside the environment to gather the necessary information that are, then, stored inside a rosbag.

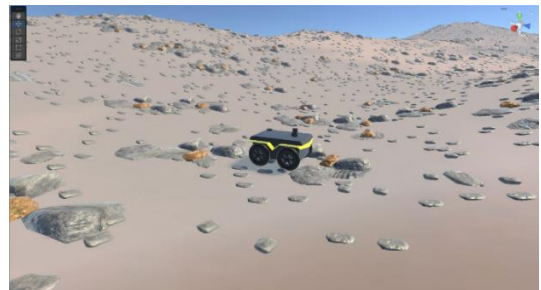

Fig 3. Unity simulation environment.

To prove the possibility of using the framework for an autonomous navigation system, the data are passed to the ros2 architecture to mirror the same operations done in an online application of the system. Step by step, the acquired data are employed in the hybrid traversability cost map's creation.

The success of these trials can be visually highlighted by verifying that optimal paths are planned and how they change as new information comes in, following the motion of the vehicle.

## 5. Results and Discussion

In this section, the results reached at the end of the testing operations and the performances exhibited by the presented framework are discussed.

Before the actual simulation tests, a preliminary analysis was carried out. It was made to check at first the usability of the cost maps as a baseline for path planning, by providing the framework with a series of images gathered through Oaisys, mirroring the rover's motion, and using them to build a kind of global cost map to test the path planning results. The outcome highlighted a behavior coherent with the expectations. The obstacles were avoided, and lower-cost areas were preferred over higher-cost ones.

The final step, having obtained a positive outcome from the preliminary tests, focuses on the use of the simulation. Fixing a goal for the rover and running the Unity environment, it has been verified that satisfactory performances are achieved.

The complete pipeline seems to work correctly, and the traversability analysis, carried out, by the proposed framework effectively employs those data to build the hybrid cost map.

An optimal path is planned over this result, the paths change their appearance as new information is considered, showing an adaptive behavior. These results demonstrate how potentially the framework could be employed in an online autonomous navigation system through simulated Martian terrain based on the commands provided by a controller to follow the planned path. As can be seen in Fig 4, obstacles are avoided and the less hazardous areas to traverse are preferred when possible.

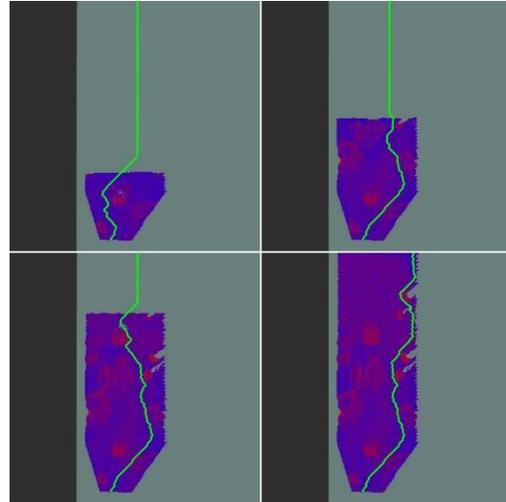

Fig 4. Path planning results after 5, 20, 35, 55 steps.

## 6. Conclusions

In this paper, a framework for terrain traversability analysis for planetary exploration rovers, particularly designed for Martian exploration, is presented. This kind of system is fundamental for autonomous navigation especially in harsh environments, like the studied one, to maintain the rover as safe as possible while trying to decouple it as much as can be done from human intervention.

A hybrid architecture is formulated involving an appearance-based and a geometry-based approach combined to yield a final cost map. The appearance-based side applies semantic segmentation and terrain's roughness classification. The geometry-based side uses, instead, terrain's slope evaluation. The two cost maps are combined to obtain the hybrid result as a 2D robot-centric grid map.

The hybrid approach has been chosen to try to increase the robustness of the framework, employing methods that can potentially compensate for each other's flaws. The weaknesses of each method, as could be the poor performances of the appearance-based approach in the presence of strong reflections, light flares, and elements hidden by shadows or the resolution's limitations of the geometry-based approach to distinguish variations below a certain scale, are balanced by the other's ability to overcome those conditions and correctly assess traversability.

The final cost map is verified to be correctly employable in an autonomous navigation system in two testing phases: a preliminary step to prove its usability for a path planning algorithm and a simulation step in which the framework assesses traversability, from the data coming from a Unity simulation, demonstrating that the rover could be able to safely and autonomously navigate in the scene reaching its target.

For what concerns possible further improvements two are the main proposals which could be considered:

- A mixed synthetic-real-world dataset could be employed in the training of the networks to obtain the same performance level reached with only synthetic data. This would possibly overcome the limitation in the development of similar systems represented by the unavailability of large real-world datasets presenting all the needed metadata and open to a possible physical implementation of the framework.
- A dedicated multitask learning model could be created to handle simultaneously semantic segmentation and roughness classification. This kind of models is proven to present improved data efficiency, reduced overfitting through shared representations, and fast learning by leveraging auxiliary information.

**References**

[1] Panagiotis Papadakis, Terrain traversability analysis methods for unmanned ground vehicles: A survey, Engineering Applications of Artificial Intelligence, Volume 26, Issue 4, 2013, Pages 1373-1385, ISSN 0952-1976, https://doi.org/10.1016/j.engappai.2013.01.006. (https://www.sciencedirect.com/science/article/pii/S095219761300016X)

[2] Rothrock, B., Kennedy, R., Cunningham, C., Papon, J., Heverly, M., & Ono, M. (2016). SPOC: Deep Learning-based Terrain Classification for Mars Rover Missions. In *AIAA SPACE Forum*. *AIAA SPACE 2016*. doi:10.2514/6.2016-5539

[3] D. Atha, R. M. Swan, A. Didier, Z. Hasnain and M. Ono, "Multi-mission Terrain Classifier for Safe Rover Navigation and Automated Science," 2022 IEEE Aerospace Conference (AERO), Big Sky, MT, USA, 2022, pp. 1-13, doi: 10.1109/AERO53065.2022.9843615. keywords: {Space vehicles;Training;Mars;Solid modeling;Adaptation models;Technological innovation;Soil properties}

[4] V. Suryamurthy, V. S. Raghavan, A. Laurenzi, N. G. Tsagarakis and D. Kanoulas, "Terrain Segmentation and Roughness Estimation using RGB Data: Path Planning Application on the CENTAURO Robot," 2019 IEEE-RAS 19th International Conference on Humanoid Robots (Humanoids), Toronto, ON, Canada, 2019, pp. 1-8, doi: 10.1109/Humanoids43949.2019.9035009. keywords: {Navigation;Mobile robots;Training;Image segmentation;Estimation;Computer architecture}

[5] Gennery, D.B. Traversability Analysis and Path Planning for a Planetary Rover. *Autonomous Robots* **6**, 131–146 (1999). https://doi.org/10.1023/A:1008831426966

[6] Joho, D., Stachniss, C., Pfaff, P., Burgard, W. (2007). Autonomous Exploration for 3D Map Learning. In: Berns, K., Luksch, T. (eds) Autonome Mobile Systeme 2007. Informatik aktuell. Springer, Berlin, Heidelberg. https://doi.org/10.1007/978-3-540-74764-2_4

[7] P. Fankhauser, M. Bloesch and M. Hutter, "Probabilistic Terrain Mapping for Mobile Robots With Uncertain Localization," in IEEE Robotics and Automation Letters, vol. 3, no. 4, pp. 3019-3026, Oct. 2018, doi: 10.1109/LRA.2018.2849506. keywords: {Robot sensing systems;Uncertainty;Legged locomotion;Position measurement;Probabilistic logic;Mapping;field robots;legged robots}

[8] Bellutta, P. & Manduchi, Roberto & Matthies, L. & Owens, K. & Rankin, A.. (2000). Terrain perception for Demo III. 326 - 331. 10.1109/IVS.2000.898363.

[9] Manduchi, Roberto & Castano, A. & Talukder, Ashit & Matthies, L.. (2005). Obstacle Detection and Terrain Classification for Autonomous Off-Road Navigation. Autonomous Robots. 18. 81-102. 10.1023/B:AURO.0000047286.62481.1d.

[10] Fusaro, Daniel & Olivastri, Emilio & Evangelista, Daniele & Iob, Pietro & Pretto, Alberto. (2022). An Hybrid Approach to Improve the Performance of Encoder-Decoder Architectures for Traversability Analysis in Urban Environments. 1745-1750. 10.1109/IV51971.2022.9827248.

[11] Leung, Tiga & Ignatyev, Dmitry & Zolotas, Argyrios. (2022). Hybrid Terrain Traversability Analysis in Off-road Environments. 50-56. 10.1109/ICARA55094.2022.9738557.

[12] R. M. Swan et al., "AI4MARS: A Dataset for Terrain-Aware Autonomous Driving on Mars," 2021 IEEE/CVF Conference on Computer Vision and Pattern Recognition Workshops (CVPRW), Nashville, TN, USA, 2021, pp. 1982-1991, doi: 10.1109/CVPRW53098.2021.00226. keywords: {Training;Space vehicles;Deep

learning;Productivity;Mars;Image segmentation;Semantics}

[13] Ronneberger, O., Fischer, P., Brox, T. (2015). U-Net: Convolutional Networks for Biomedical Image Segmentation. In: Navab, N., Hornegger, J., Wells, W., Frangi, A. (eds) Medical Image Computing and Computer-Assisted Intervention – MICCAI 2015. MICCAI 2015. Lecture Notes in Computer Science(), vol 9351. Springer, Cham. https://doi.org/10.1007/978-3-319-24574-4_28

[14] Kingma, Diederik & Ba, Jimmy. (2014). Adam: A Method for Stochastic Optimization. International Conference on Learning Representations.

[15] Zhou, Qian-Yi & Park, Jaesik & Koltun, Vladlen. (2018). Open3D: A Modern Library for 3D Data Processing.

[16] M. G. Müller, M. Durner, A. Gawel, W. Stürzl, R. Triebel and R. Siegwart, "A Photorealistic Terrain Simulation Pipeline for Unstructured Outdoor Environments," 2021 IEEE/RSJ International Conference on Intelligent Robots and Systems (IROS), Prague, Czech Republic, 2021, pp. 9765-9772, doi: 10.1109/IROS51168.2021.9636644.

[17] https://github.com/Unity-Technologies/ROS-TCP-Endpoint